\documentclass[runningheads]{llncs}
\usepackage{graphicx}
\usepackage{floatrow}
\newfloatcommand{capbtabbox}{table}[][\FBwidth]
\usepackage{blindtext}
\usepackage{wrapfig}
\usepackage{times}
\usepackage{xcolor}
\usepackage{soul}
\usepackage[utf8]{inputenc}
\usepackage{hyperref}

\usepackage[normalem]{ulem}
\usepackage{epsfig}
\newcommand{\tht}[2]{\begin{tabular}{@{}#1@{}}#2\end{tabular}}
\usepackage{bm,algorithm,algpseudocode,threeparttable,subfigure,rotating,multirow,array,graphics,textcomp}
\usepackage{setspace}
\usepackage{booktabs}
\usepackage{amsmath}
\usepackage{amssymb}
\graphicspath{{Figures/}}
\usepackage{enumitem}
\usepackage{wrapfig}

\def\eg{\textit{e.g.}}
\def\ie{\textit{i.e.}}

\def\etal{\textit{et al.}}

\begin{document}

\title{Dual Generator Generative Adversarial Networks for Multi-Domain Image-to-Image Translation
	\thanks{This work was supported by NIST 60NANB17D191. We gratefully acknowledge the gift donations of Cisco, Inc. and the support of NVIDIA Corporation with the donation of the GPUs used for this research. This article solely reflects the opinions and conclusions of its authors and neither NIST, Cisco, nor NVIDIA.}
}
\titlerunning{G$^2$GAN for Multi-Domain Image-to-Image Translation}

\author{Hao Tang\inst{1}\and
	Dan Xu\inst{2}\and
	Wei Wang\inst{3}\and
	Yan Yan\inst{4}\and
	Nicu Sebe\inst{1}}

\authorrunning{Hao Tang, Dan Xu, Wei Wang, Yan Yan, Nicu Sebe} 

\institute{University of Trento, Italy \and
	University of Oxford, United Kingdom \and
	\'Ecole Polytechnique F\'ed\'erale de Lausanne, Switzerland \and
	Texas State University, USA
}

\maketitle


\begin{abstract}
	
	State-of-the-art methods for image-to-image translation with Generative Adversarial Networks (GANs) can learn a mapping from one domain to another domain using unpaired image data. However, these methods require the training of one specific model for every pair of image domains, which limits the scalability in dealing with more than two image domains.
	In addition, the training stage of these methods has the common problem of model collapse that degrades the quality of the generated images. 
	To tackle these issues, we propose a Dual Generator Generative Adversarial Network (G$^2$GAN), which is a robust and scalable approach allowing to perform unpaired image-to-image translation for multiple domains using only dual generators within a single model.
	Moreover, we explore different optimization losses for better training of G$^2$GAN, and thus make unpaired image-to-image translation with higher consistency and better stability. 
	Extensive experiments on six publicly available datasets with different scenarios, \textit{i.e.}, architectural buildings, seasons, landscape and human faces, demonstrate that the proposed G$^2$GAN achieves superior model capacity and better generation performance comparing with existing image-to-image translation GAN models.
	
\end{abstract}

\section{Introduction}
Generative Adversarial Networks (GANs)~\cite{goodfellow2014generative} have recently received considerable attention in various communities, \emph{e.g.}, computer vision, natural language processing and medical analysis. GANs are generative models which are particularly designed for image generation tasks.
Recent works have been able to yield promising image-to-image translation performance (\emph{e.g.}, pix2pix \cite{isola2017image} and BicycleGAN \cite{zhu2017toward}) in a supervised setting given carefully annotated image pairs. 
However, pairing the training data is usually difficult and costly. The situation becomes even worse when dealing with tasks such as artistic stylization, since the desired output is very complex, typically requiring artistic authoring. 
To tackle this problem, several GAN approaches, such as CycleGAN~\cite{zhu2017unpaired}, DualGAN \cite{yi2017dualgan}, DiscoGAN \cite{kim2017learning}, ComboGAN \cite{anoosheh2017combogan} and DistanceGAN \cite{benaim2017one}, aim to effectively learn a hidden mapping from one image domain to another image domain with unpaired image data. However, these cross-modal translation frameworks are not efficient for multi-domain image-to-image translation. 
For instance, given $m$ image domains, pix2pix  and BicycleGAN require the training of $A_m^2{=}m(m{-}1){=}\mathrm{\Theta}(m^2)$ models; CycleGAN, DiscoGAN, DistanceGAN and DualGAN need $C_m^2{=}\frac{m(m{-}1)}{2}{=}\mathrm{\Theta}(m^2)$ models or $m(m{-}1)$ generator/discriminator pairs; ComboGAN requires $\mathrm{\Theta}(m)$ models.

\begin{figure}[!t] \tiny
	\centering
	\includegraphics[width=1\linewidth]{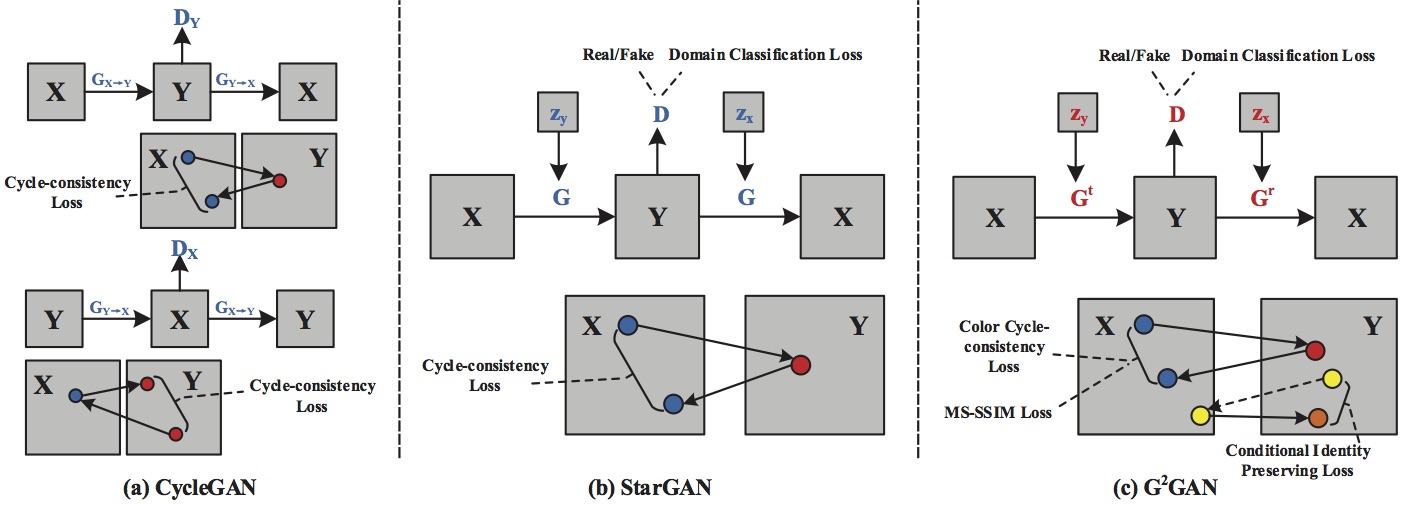}
	\caption{A motivation illustration of the proposed G$^2$GAN (c) compared with CycleGAN (a) \protect\cite{zhu2017unpaired}  and StarGAN~\cite{choi2017stargan}  (b). For the multi-domain image-to-image generation task w.r.t. $m$ image domains, CycleGAN needs to train $m(m{-}1)$ generator/discriminator pairs, while the proposed G$^2$GAN only needs to train dual generators and one discriminator. StarGAN~\cite{choi2017stargan} shares the same generator for both the translation and reconstruction tasks, while G$^2$GAN employs task-specific generators ($G^t$ and $G^r$) which allow for different network designs and different levels of parameter sharing.}
	\label{fig:motivation}
\end{figure}

To overcome the aforementioned limitation, Choi~\etal~propose StarGAN \cite{choi2017stargan} (Figure~\ref{fig:motivation}(b)), which can perform multi-domain image-to-image translation using only one generator/discriminator pair with the aid of an auxiliary classifier \cite{odena2016conditional}. More formally, let $X$ and $Y$ represent training sets of two different domains, and $x{\in} X$ and $y{\in} Y$ denote training images in domain $X$ and domain $Y$, respectively; let $z_y$ and $z_x$ indicate category labels of domain $Y$ and $X$, respectively.
StarGAN utilizes the same generator $G$ twice for translating from $X$  to $Y$ with the labels $z_y$, \ie, $G(x, z_y){\approx} y$, and reconstructs the input $x$ from the translated output $G(x, z_y)$ and the label $z_x$, \ie, $G(G(x, z_y), z_x){\approx} x$. In doing so, the same generator shares a common mapping and data structures for two different tasks, \ie, translation and reconstruction. However, since each task has unique information and distinct targets, it is harder to optimize the generator and to make it gain good generalization ability on both tasks, which usually leads to blurred generation results. 

In this paper, we propose a novel Dual Generator Generative Adversarial Network (G$^2$GAN) (Figure~\ref{fig:motivation}(c)). Unlike StarGAN, G$^2$GAN consists of two generators and one discriminator, the  translation generator $G^t$ transforms images from $X$ to $Y$, and the reconstruction generator $G^r$ uses the generated images from $G^t$ and the original domain label $z_x$ to reconstruct the original $x$. 
Generators $G_t$ and $G_r$ cope with different tasks, and the input data distribution for them is different. 
The input of $G_t$ is a real image and a target domain label. 
The goal of $G_t$ is to generate the target domain image. 
While $G_r$ accepts a generated image and an original domain label as input, the goal of $G_r$ is to generate an original image. 
For $G_t$ and $G_r$, the input images are a real image and a generated image, respectively.
Therefore, it is intuitive to design different network structures for the two generators.
The two generators are allowed to use different network designs and different levels of parameter sharing according to the diverse difficulty of the tasks. In this way, each generator can have its own network parts which usually helps to learn better each task-specific mapping in a multi-task setting~\cite{ruder2017overview}. 

To overcome the model collapse issue in training G$^2$GAN for the multi-domain translation, we further explore different objective losses for better optimization. The proposed losses include (i) a color cycle-consistency loss which targets solving the ``channel pollution'' problem  \cite{tang2018gesturegan}  by generating red, green, blue channels separately instead of generating all three at one time, (ii) a multi-scale SSIM loss, which preserves the information of luminance, contrast and structure between generated images and input images across different scales, and (iii) a conditional identity preserving loss, which helps retaining the identity information of the input images. These losses are jointly embedded in G$^2$GAN for training and help generating results with higher consistency and better stability. In summary, the contributions of this paper are as follows: 
\begin{itemize}
	\renewcommand{\labelitemi}{$\bullet$}
	\item We propose a novel Dual Generator Generative Adversarial Network (G$^2$GAN), which can perform unpaired image-to-image translation among multiple image domains. The dual generators, allowing different network structures and different-level parameter sharing, are designed to specifically cope with the translation and the reconstruction tasks, which facilitates obtaining a better generalization ability of the model to improve the generation quality.
	\item We explore jointly utilizing different objectives for a better optimization of the proposed G$^2$GAN, and thus obtaining unpaired multi-modality translation with higher consistency and better stability.
	\item We extensively evaluate G$^2$GAN on six different datasets in different scenarios, such as architectural buildings, seasons, landscape and human faces, demonstrating its superiority in model capacity and its better generation performance compared with state-of-the-art methods on the multi-domain image-to-image translation task. 
\end{itemize}

\section{Related Work}

\noindent \textbf{Generative Adversarial Networks (GANs)} \cite{goodfellow2014generative} are powerful generative models, which have achieved impressive results on different computer vision tasks, \emph{e.g.},
image generation \cite{shrivastava2016learning,park2017transformation}, editing \cite{brock2016neural,shu2017neural} and inpainting \cite{li2017generative,yeh2016semantic}. 
However, GANs are difficult to train, since it is hard to keep the balance between the generator and the discriminator, which makes the optimization oscillate and thus leading to a collapse of the generator. To address this, several solutions have been proposed recently, such as Wasserstein GAN \cite{arjovsky2017wasserstein} and Loss-Sensitive GAN~\cite{qi2017loss}. 
To generate more meaningful images, CGAN~\cite{mirza2014conditional} has been proposed to employ conditioned information to guide the image generation. 
Extra information can also be used such as discrete category labels \cite{perarnau2016invertible,liang2017generative}, text descriptions \cite{mansimov2015generating,reed2016generative}, object/face keypoints \cite{reed2016learning,wang2018every}, human skeleton \cite{tang2018gesturegan,siarohin2018deformable} and referenced images \cite{isola2017image,li2018beautygan}. CGAN models have been successfully used in various applications, such as image editing \cite{perarnau2016invertible}, text-to-image translation~\cite{mansimov2015generating} and image-to-image translation \cite{isola2017image} tasks. 

\noindent \textbf{Image-to-Image Translation.} CGAN models learn a translation between image inputs and image outputs using neutral networks. 
Isola~\etal~\cite{isola2017image} design the pix2pix framework which is a conditional framework using a CGAN to learn the mapping function. 
Based on pix2pix, Zhu~\etal~\cite{zhu2017toward} further present BicycleGAN which achieves multi-modal image-to-image translation using paired data. 
Similar ideas have also been applied to many other tasks, \eg~generating photographs from sketches \cite{sangkloy2016scribbler}. However, most of the models require paired training data, which are usually costly to obtain.

\noindent \textbf{Unpaired Image-to-Image Translation.} To alleviate the issue of pairing training data, Zhu~\etal~\cite{zhu2017unpaired} introduce CycleGAN, which learns the mappings between two unpaired image domains without supervision with the aid of a cycle-consistency loss. 
Apart from CycleGAN, there are other variants proposed to tackle the problem. 
For instance, CoupledGAN~\cite{liu2016coupled} 
uses a weight-sharing strategy to learn common representations across domains. 
Taigman~\etal~\cite{taigman2016unsupervised} propose a Domain Transfer Network (DTN) which learns a generative function between one domain and another domain. 
Liu~\etal~\cite{liu2017unsupervised} extend the basic structure of GANs via combining the Variational Autoencoders (VAEs) and GANs. 
A novel DualGAN mechanism is demonstrated in~\cite{yi2017dualgan}, in which image translators are trained from two unlabeled image sets each representing an image domain. Kim~\etal~\cite{kim2017learning} propose a method based on GANs that learns to discover relations between different domains. 
However, these models are only suitable in cross-domain translation problems.

\noindent \textbf{Multi-Domain Unpaired Image-to-Image Translation.} There are only very few recent methods attempting to implement multi-modal image-to-image translation in an efficient way. Anoosheh~\etal~propose a ComboGAN model~\cite{anoosheh2017combogan}, which only needs to train $m$ generator/discriminator pairs for $m$ different image domains. To further reduce the model complexity, Choi~\etal~introduce StarGAN \cite{choi2017stargan}, which has a single generator/discriminator pair and is able to perform the task with a complexity of $\mathrm{\Theta}(1)$. Although the model complexity is low, jointly learning both the translation and reconstruction tasks with the same generator requires the sharing of all parameters, which increases the optimization complexity and reduces the generalization ability, thus leading to unsatisfactory generation performance. The proposed approach aims at obtaining a good balance between the model capacity and the generation quality. Along this research line, we propose a Dual Generator Generative Adversarial Network (G$^2$GAN), which 
achieves this target via using two task-specific generators and one discriminator. We also explore various optimization objectives to train better the model to produce more consistent and more stable results.

\begin{figure}[!t] \tiny
	\centering
	\includegraphics[width=1\linewidth]{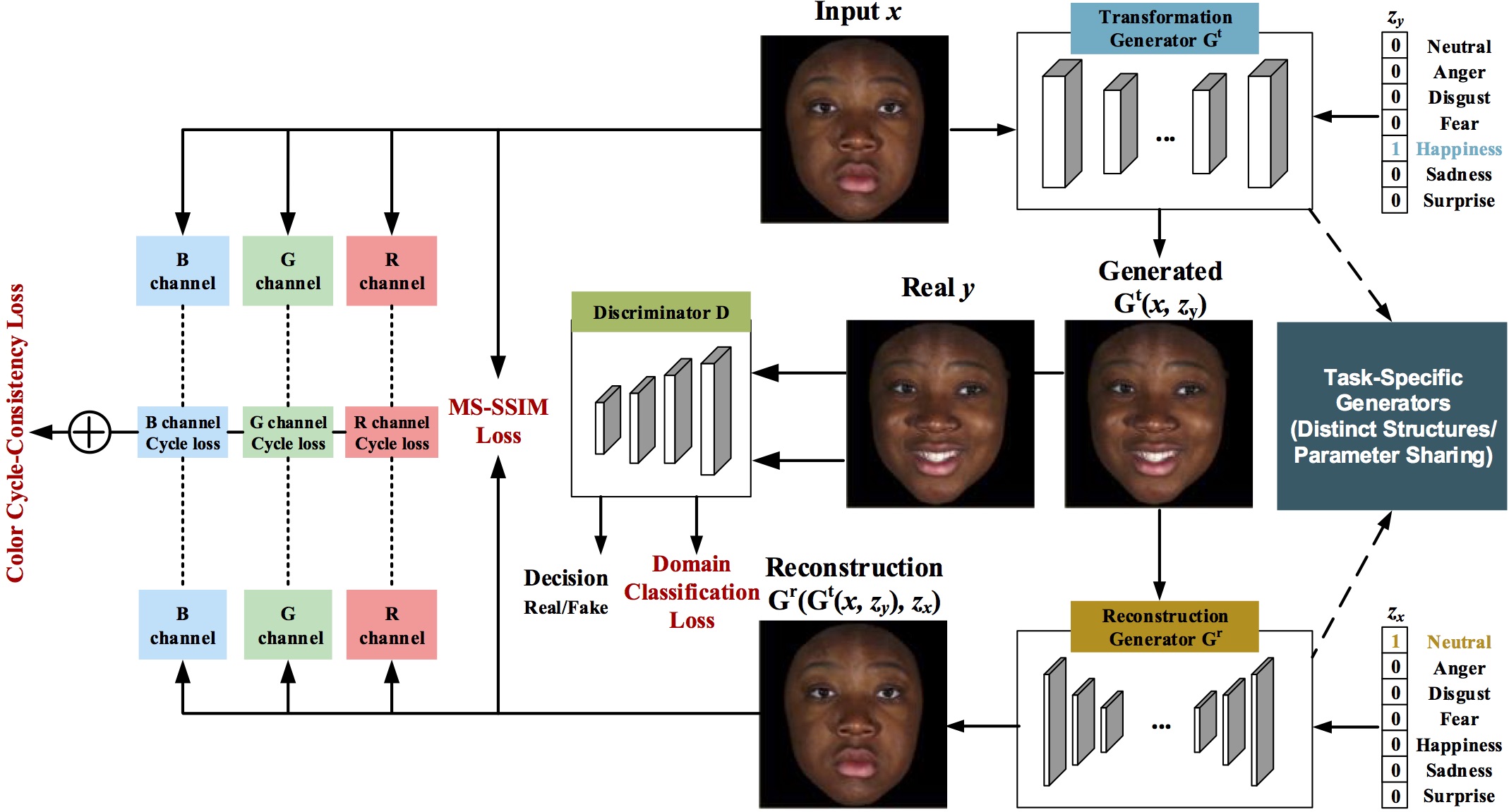}
	\caption{The framework of G$^2$GAN. $z_x$ and $z_y$ indicate the category labels of domain $X$ and $Y$, respectively. $G^t$ and $G^r$ are task-specific generators. The generator $G^t$ converts images from domain $X$ into domain $Y$ and the generator $G^r$ inputs the generated image $G^t(x, z_y)$ and the original domain label $z_x$ and attempts to reconstruct the original image $x$ during the optimization with the proposed different objective losses.}
	\label{fig:mainFramework}
\end{figure}

\section{G$^2$GAN: Dual Generator Generative Adversary Networks}
We first start with the model formulation of G$^2$GAN, and then introduce the proposed objectives for better optimization of the model, and finally present the implementation details of the whole model including network architecture and training procedure.
\subsection{Model Formulation}
In this work, we focus on the multi-domain image-to-image translation task with unpaired training data. The overview of the proposed G$^2$GAN is depicted in Figure~\ref{fig:mainFramework}. 
The proposed G$^2$GAN model is specifically designed for tackling the multi-domain translation problem with significant advantages in the model complexity and in the training overhead compared with the cross-domain generation models, such as CycleGAN~\cite{zhu2017unpaired}, DiscoGAN \cite{kim2017learning} and DualGAN \cite{yi2017dualgan}, which need to separately train $C_m^2{=}\frac{m(m-1)}{2}$ models for $m$ different image domains, while ours only needs to train a single model.
To directly compare with StarGAN \cite{choi2017stargan}, which simply employs the same generator for the different reconstruction and translation tasks.
However, the training of a single generator model for multiple domains is a challenging problem (refer to Section \ref{sec:exp}), we proposed a more effective dual generator network structure and more robust optimization objectives to stabilize the training process. Our work focuses on exploring different strategies to improve the optimization of the multi-domain model aiming to give useful insights in the design of more effective multi-domain generators.

Our goal is to learn all the mappings among multiple domains using dual generators and one discriminator. To achieve this target, we train a translation generator $G^t$ to convert an input image $x$ into an output image $y$ which is conditioned on the target domain label $z_y$, \ie~$G^t(x, z_y){\rightarrow}y$. Then the reconstruction generator $G^r$ accepts the generated image $G^t(x, z_y)$ and the original domain label $z_x$ as input, and learns to reconstruct the input image $x$, \ie~$G^r(G^t(x, z_y), z_x){\rightarrow}x$ through the proposed different optimization losses, including the color cycle-consistency loss for solving the ``channel pollution'' issue and the MS-SSIM loss for preserving the information of luminance, contrast and structure across scales. The dual generators are task-specific generators which allows for different network designs and different levels of parameter sharing for learning better the generators. The discriminator $D$ tries to distinguish between the real image $y$ and the generated image $G^t(x, z_y)$, and to classify the generated image $G^t(x, z_y)$ to the target domain label $z_y$ via the domain classification loss.  
We further investigate how the distinct network designs and different network sharing schemes for the dual generators dealing with different sub-tasks could balance the generation performance and the network complexity. The multi-domain model StarGAN \cite{choi2017stargan} did not consider these aspects. 

\subsection{Model Optimization}
The optimization objective of the proposed G$^2$GAN contains five different losses, \textit{i.e.}, color cycle-consistency loss, multi-scale SSIM loss, conditional least square loss, domain classification loss and conditional identity preserving loss. These optimization losses are jointly embedded into the model during training. We present the details of these loss functions in the following.

\noindent\textbf{Color Cycle-Consistency Loss.}
It is worth noting that CycleGAN \cite{zhu2017unpaired} is different from the pix2pix framework \cite{isola2017image} as the training data in CycleGAN are unpaired, and thus CycleGAN introduces a cycle-consistency loss to enforce forward-backward consistency. 
The core idea of ``cycle consistency'' is that if we translate from one domain to the other and translate back again we should arrive at where we started. 
This loss can be regarded as ``pseudo'' pairs in training data even though we do not have corresponding samples in the target domain for the input data in the source domain. Thus, the loss function of cycle-consistency is defined as:

\begin{equation} \small
\begin{aligned}
& \mathcal{L}_{cyc}(G^t, G^r, x, z_x, z_y) 
=  \mathbb{E}_{x\sim{p_{\rm data}}(x)}[\Arrowvert G^r(G^t(x, z_y), z_x)-x\Arrowvert_1].
\end{aligned}
\label{equ:cycleganloss}
\end{equation}
The optimization objective is to make the reconstructed images $G^r(G^t(x, z_y), z_x)$ as close as possible to the input images $x$, and the $L_1$ norm is adopted for the reconstruction loss.
However, the ``channel pollution'' issue  \cite{tang2018gesturegan}  exists in this loss, which is because the generation of a whole image at one time makes the different channels influence each other, thus leading to artifacts in the generation results. To solve this issue, we propose to construct the consistence loss for each channel separately, and introduce the color cycle-consistency loss as follows:

\begin{equation} \small
\begin{aligned}
& \mathcal{L}_{colorcyc}(G^t, G^r, x, z_x, z_y) = \sum_{i\in\{r,g,b\}}\mathcal{L}_{cyc}^{i}(G^t, G^r, x^i, z_x, z_y),
\end{aligned}
\label{equ:colorcycleganloss}
\end{equation}
where, ${x^b, x^g, x^r}$ are three color channels of image $x$.
Note that we did not feed each channel of the image into the generator separately.
Instead, we feed the whole image into the generator. We calculate the pixel loss for the red, green, blue channels separately between the reconstructed image and the input image, and then sum up the three distance losses as the final loss. By doing so, the generator can be enforced to generate each channel independently to avoid the ``channel pollution'' issue. 

\noindent\textbf{Multi-Scale SSIM Loss.}
The structural similarity index (SSIM) has been originally used in~\cite{wang2004image} to measure the similarity of two images. We introduce it into the proposed G$^2$GAN to help preserving the information of luminance, contrast and structure across scales. For the recovered image $\widehat{x}{=}G^r(G^t(x,z_y),z_x)$ and the input image $x$, the SSIM loss is written as:
\begin{equation} \small
\begin{aligned}
\mathcal{L}_{\rm{SSIM}}(\widehat{x},x)=\left[l(\widehat{x},x)\right]^\alpha \left[c(\widehat{x}, x)\right]^\beta \left[s(\widehat{x},x) \right]^\gamma,
\end{aligned}
\end{equation}
where 
\begin{equation} \small
\begin{aligned}
l(\widehat{x},x)=\frac{2\mu_{\widehat{x}}\mu_x + C_1}{\mu_{\widehat{x}}^2 +\mu_x^2 +C_1}, \quad c(\widehat{x},x)=\frac{2\sigma_{\widehat{x}}\sigma_x + C_2}{\sigma_{\widehat{x}}^2 + \sigma_x^2 + C_2}, \quad s(\widehat{x},x)=\frac{\sigma_{{\widehat{x}}x}+C_3}{\sigma_{\widehat{x}}\sigma_x+C_3}.
\label{equ:msssim}
\end{aligned}
\end{equation} 
These three terms compare the luminance, contrast and structure information between $\widehat{x}$ and $x$ respectively. The parameters $\alpha{>}0$, $\beta{>}0$ and $\gamma{>}0$ control the relative importance of the $l(\widehat{x},x)$, $c(\widehat{x},x)$ and $s(\widehat{x},x)$, respectively; $\mu_{\widehat{x}}$ and $\mu_x$ are the means of $\widehat{x}$ and $x$; $\sigma_{\widehat{x}}$ and $\sigma_x$ are the standard deviations of $\widehat{x}$ and $x$; $\sigma_{\widehat{x}x}$ is the covariance of $\widehat{x}$ and $x$; $C_1$, $C_2$ and $C_3$ are predefined parameters. To make the model benefit from multi-scale deep information, we refer to a multi-scale implementation of SSIM~\cite{wang2003multiscale} which constrains SSIM over scales. We write the Multi-Scale SSIM (MS-SSIM) as:
\begin{equation} \small
\mathcal{L}_{\rm{MS-SSIM}}(\widehat{x},x) = \left[l_M(\widehat{x},x)\right]^{\alpha_M} \prod\limits_{j=1}^M \left[ c_j(\widehat{x},x)\right]^{\beta_j} \left[s_j(\widehat{x},x)\right]^{\gamma_j}.
\label{equ:ssim}
\end{equation}
Through using the MS-SSIM loss, the luminance, contrast and structure information of the input images is expected to be preserved.

\noindent\textbf{Conditional Least Square Loss.}  We apply a least square loss \cite{mao2017least,zhu2017unpaired} to stabilize our model during training.
The least square loss is more stable than the negative log likelihood objective $ \mathcal{L}_{CGAN}(G^t, D_s, z_y) = \mathbb{E}_{y\sim{p_{\rm data}}(y)}\left[ \log D_s(y)\right] + \mathbb{E}_{x\sim{p_{\rm data}}(x)}[\log (1 - D_s(G^t(x, z_y)))]$, and is converging faster than the Wasserstein GAN (WGAN)~\cite{arjovsky2017wasserstein}. The loss can be expressed as:
\begin{equation} \small
\begin{aligned}
\mathcal{L}_{LSGAN}(G^t, D_s, z_y) =
\mathbb{E}_{y\sim{p_{\rm data}(y)}}[(D_s(y)-1)^2] 
+ \mathbb{E}_{x\sim{p_{\rm data}}(x)}[D_s( G^t(x, z_y))^2],
\end{aligned}
\label{equ:legan}
\end{equation}
where $z_y$ are the category labels of domain $y$, $D_s$ is the probability distribution over sources produced by discriminator $D$. 
The target of $G^t$ is to generate an image $G^t(x,z_y)$ that is expected to be similar to the images from domain $Y$, while $D$ aims to distinguish between the generated images $G^t(x,z_y)$ and the real images $y$.

\noindent\textbf{Domain Classification Loss.}
To perform multi-domain image translation with a single discriminator, previous works employ an auxiliary classifier~\cite{odena2016conditional,choi2017stargan} on the top of discriminator, and impose the domain classification loss when updating both the generator and discriminator. We also consider this loss in our optimization:
\begin{equation} \small
\begin{aligned}
\mathcal{L}_{classification}(G^t, D_c,z_x,z_y) = 
\mathbb{E}_{x\sim{p_{\rm data}(x)}}\{-[\log D_c(z_x|x) + \log D_c(z_y|G^t(x, z_y))]\},
\end{aligned}
\label{equ:class}
\end{equation}
where $D_c(z_x|x)$ represents the probability distribution over the domain labels given by discriminator $D$. 
$D$ learns to classify $x$ to its corresponding domain $z_x$. 
$D_c(z_y|G^t(x, z_y)$ denotes the domain classification for fake images. We minimize this objective function to generate images $G^t(x, z_y)$ that can be classified as the target labels $z_y$. 

\noindent\textbf{Conditional Identity Preserving Loss.}
To reinforce the identity of the input image during conversion, a conditional identity preserving loss \cite{zhu2017unpaired,taigman2016unsupervised}  is used.
This loss can encourage the mapping to preserve color composition between the input and the output, and can regularize the generator to be near an identity mapping when real images of the target domain are provided as the input to the generator. 
\begin{equation}\small
\begin{aligned}
& \mathcal{L}_{identity}(G^t, G^r, z_x)= \mathbb{E}_{x\sim{p_{\rm data}}(x)}[\Arrowvert G^r(x,z_x)-x\Arrowvert_1].
\end{aligned}
\end{equation}
In this way, the generator also takes into account the identity preserving via the back-propagation of the identity loss.
Without this loss, the generators are free to change the tint of input images when there is no need to.

\noindent\textbf{Full G$^2$GAN Objective.} Given the losses presented above, the complete optimization objective of the proposed G$^2$GAN can be written as:
\begin{equation} \small
\begin{aligned}
& \mathcal{L} = \mathcal{L}_{LSGAN}+ \lambda_1\mathcal{L}_{classification} +
\lambda_2 \mathcal{L}_{colorcyc} + \lambda_3 \mathcal{L}_{\rm{MS-SSIM}} + \lambda_4 \mathcal{L}_{identity},
\end{aligned}
\label{eqn:allloss}
\end{equation}
where $\lambda_1$, $\lambda_2$, $\lambda_3$ and $\lambda_4$ are parameters controlling the relative importance of the corresponding objectives terms. All objectives are jointly optimized in an end-to-end fashion.

\subsection{Implementation Details}
\noindent\textbf{G$^2$GAN Architecture.} 
The network consists of a dual generator and a discriminator. The dual generator is designed to specifically deal with different tasks in GANs, \ie~the translation and the reconstruction tasks, which has different targets for training the network. We can design different network structures for the different generators to make them learn better task-specific objectives. This also allows us to share parameters between the generators to further reduce the model capacity, since the shallow image representations are sharable for both generators. The parameter sharing facilitates the achievement of good balance between the model complexity and the generation quality. Our model generalizes the model of StarGAN~\cite{choi2017stargan}. When the parameters are fully shared with the usage of the same network structure for both generators, our basic structure becomes a StarGAN. For the discriminator, we employ PatchGAN~\cite{isola2017image,zhu2017unpaired,li2016precomputed,choi2017stargan}. After the discriminator, a convolution layer is applied to produce a final one-dimensional output which indicates whether local image patches are real or fake. 

\noindent\textbf{Network Training.} 
For reducing model oscillation, we adopt the strategy in~\cite{shrivastava2017learning} which uses a cache of generated images to update the discriminator. In the experiments, we set the number of image buffer to 50. 
We employ the Adam optimizer \cite{kingma2014adam} to optimize the whole model.
We sequentially update the translation generator and the reconstruction generator after the discriminator updates at each iteration. The batch size is set to 1 for all the experiments and all the models were trained with 200 epochs. We keep the same learning rate for the first 100 epochs and linearly decay the rate to zero during the next 100 epochs. Weights were initialized from a Gaussian distribution with mean 0 and standard deviation 0.02.

\section{Experiments}
\label{sec:exp}
In this section, we first introduce the experimental setup, and then show detailed qualitative and quantitative results and model analysis.
\subsection{Experimental Setup}
\label{expsetup}
\begin{table*}[!t] \tiny
	\centering
	\caption{Description of the datasets used in our experiments.}
	\resizebox{0.95\linewidth}{!}{
		\begin{tabular}{l|c|c|c|c|c|c|c|c} \toprule		    
			Dataset                                                         & Type                    & \# Domain & \# Translation & Resolution            & Unpaired/Paired & \# Training   & \# Testing   & \# Total \\ \midrule
			\textbf{Facades} \cite{tylevcek2013spatial}    & Architectures      & 2               & 2                    & 256$\times$256    & Paired               & 800           & 212          & 1,012 \\ \hline
			\textbf{AR} \cite{martinez1998ar}                   & Faces                  & 4              & 12                   & 768$\times$576    & Paired                & 920           & 100          & 1,020 \\ \hline
			\textbf{Bu3dfe} \cite{yin20063d}                   & Faces                  & 7               & 42                   & 512$\times$512    & Paired               & 2,520         & 280          & 2,800 \\ \hline
			\textbf{Alps} \cite{anoosheh2017combogan} & Natural Seasons & 4                & 12                   & -                           & Unpaired          & 6,053         & 400          & 6,453 \\ \hline
			\textbf{RaFD} \cite{langner2010presentation} & Faces                  & 8               & 56                  & 1024$\times$681   & Unpaired          &  5,360       & 2,680       & 8,040 \\ \hline
			\textbf{Collection} \cite{zhu2017unpaired}     & Painting Style      & 5               & 20	                & 256$\times$256     & Unpaired          & 7,837         & 1,593        & 9,430 \\	
			\bottomrule		
	\end{tabular}}
	\label{tab:dataset}
\end{table*}

\noindent\textbf{Datasets.}
We employ six publicly available datasets to validate our G$^2$GAN. A detailed comparison of these datasets is shown in Table~\ref{tab:dataset}, including Facades, AR Face, Alps Season, Bu3dfe, RaFD and Collection style datasets.

\noindent\textbf{Parameter Setting.}
The initial learning rate for Adam optimizer is 0.0002, and $\beta_1$ and $\beta_2$ of Adam are set to 0.5 and 0.999. 
The parameters $\lambda_1, \lambda_2, \lambda_3, \lambda_4$ in Equation \ref{eqn:allloss} are set to 1, 10, 1, 0.5, respectively. 
The parameters $C_1$ and $C_2$ in Equation~ \ref{equ:msssim} are set to $0.01^2$ and $0.03^3$. The proposed G$^2$GAN is implemented using deep learning framework PyTorch. 
Experiments are conducted on an NVIDIA TITAN Xp GPU. 

\noindent\textbf{Baseline Models.} We consider several state-of-the-art cross-domain image generation models, \ie~CycleGAN~\cite{zhu2017unpaired}, DistanceGAN \cite{benaim2017one}, Dist. + Cycle \cite{benaim2017one}, Self Dist. \cite{benaim2017one}, DualGAN~\cite{yi2017dualgan}, ComboGAN \cite{anoosheh2017combogan}, BicycleGAN \cite{zhu2017toward}, pix2pix~\cite{isola2017image} as our baselines. For comparison, we train these models multiple times for every pair of two different image domains except for ComboGAN \cite{anoosheh2017combogan}, which needs to train $m$ models for $m$ different domains.
We also employ StarGAN~\cite{choi2017stargan} as a baseline which can perform multi-domain image translation using one generator/discriminator pair. 
Note that the fully supervised pix2pix and BicycleGAN are trained on paired data, the other baselines and G$^2$GAN are trained with unpaired data.
Since BicycleGAN can generate several different outputs with one single input image, and we randomly select one output from them for comparison.
For a fair comparison, we re-implement baselines using the same training strategy as our approach. 

\subsection{Comparison with the State-of-the-Art on Different Tasks}
We evaluate the proposed G$^2$GAN on four different tasks, \textit{i.e.}, label$\leftrightarrow$photo translation, facial expression synthesis, season translation and painting style transfer.  The comparison with the state-of-the-arts are described in the following.

\begin{figure}[!t] \tiny
	\centering
	\includegraphics[width=0.76\linewidth]{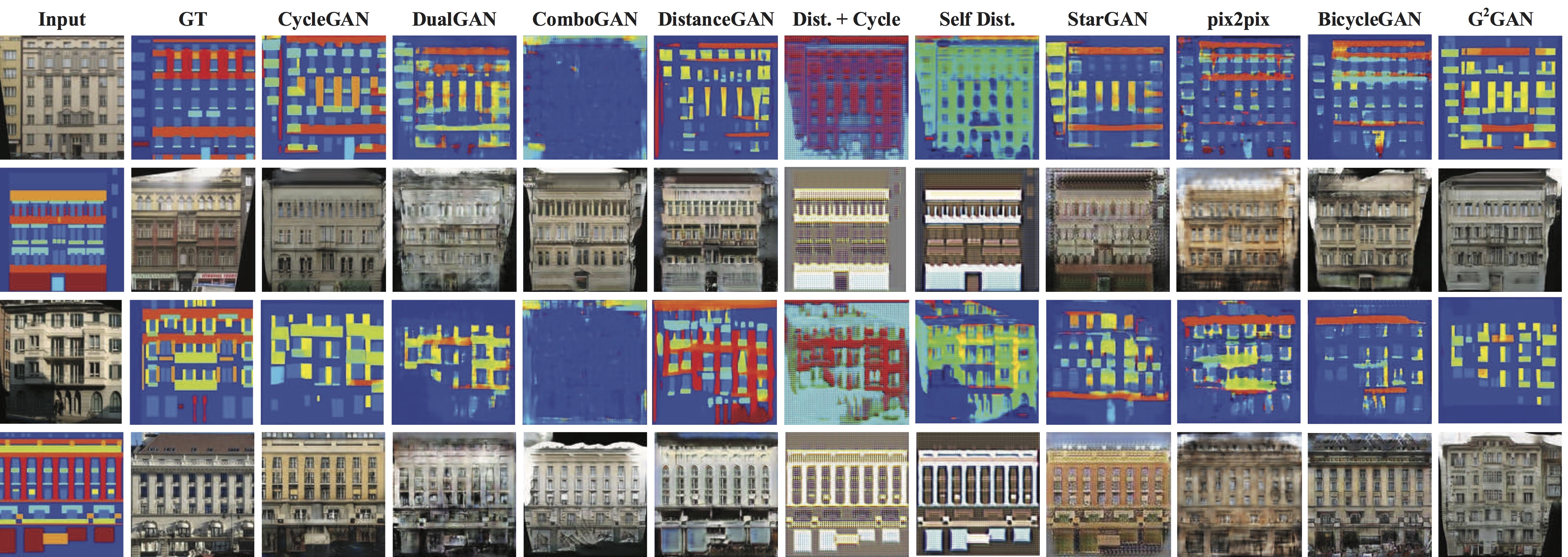}
	\caption{Comparison with different models for mapping label$\leftrightarrow$photo on Facades.
	}
	\label{fig:comparison_facades}
\end{figure}

\noindent\textbf{Task 1: Label$\leftrightarrow$Photo Translation.}
We employ Facades dataset for the label$\leftrightarrow$photo translation. 
The results on Facades were only meant to show that the proposed model is also applicable on translation on two domains only and could produce competitive performance.
The qualitative comparison is shown in Figure \ref{fig:comparison_facades}. We can obverse that ComboGAN, Dist. + Cycle, Self Dist. fail to generate reasonable results on the photo to label translation task. For the opposite mapping, \ie~(labels$\rightarrow$photos), DualGAN, Dist. + Cycle, Self Dist., StarGAN and pix2pix suffer from the model collapse problem, which leads to reasonable but blurry generation results. The proposed G$^2$GAN achieves compelling results on both tasks compared with the other baselines. 

\begin{figure}[!t] \tiny
	\centering
	\includegraphics[width=0.76\linewidth]{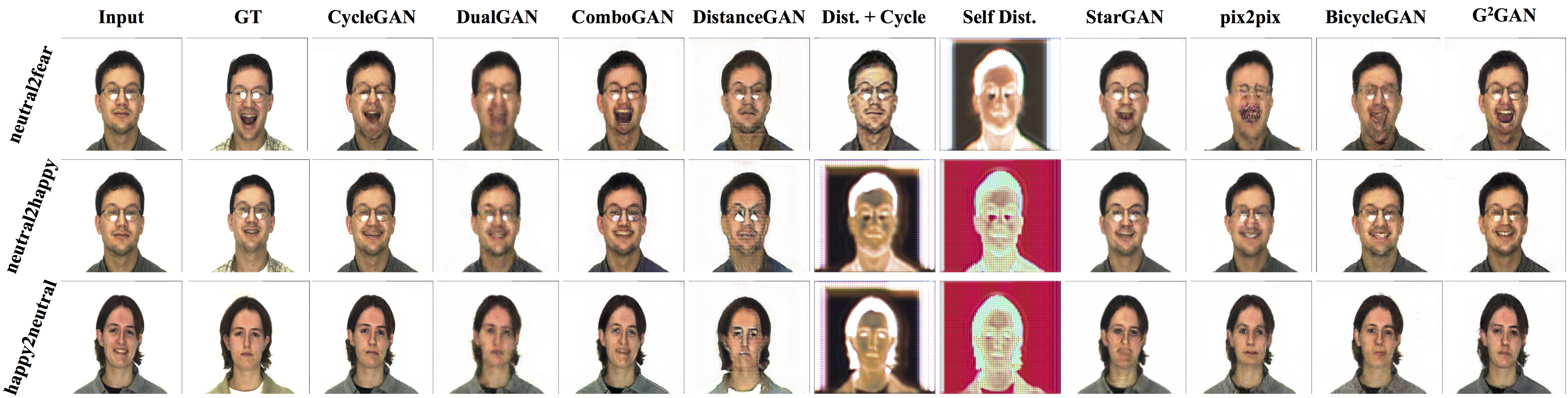}
	\caption{Comparison with different models for facial expression translation on AR. 
	}
	\label{fig:comparison_ar}
\end{figure}

\begin{figure*}[!t]
	\centering
	\setcounter{subfigure}{0}
	\subfigure{\includegraphics[width=0.59\textwidth]{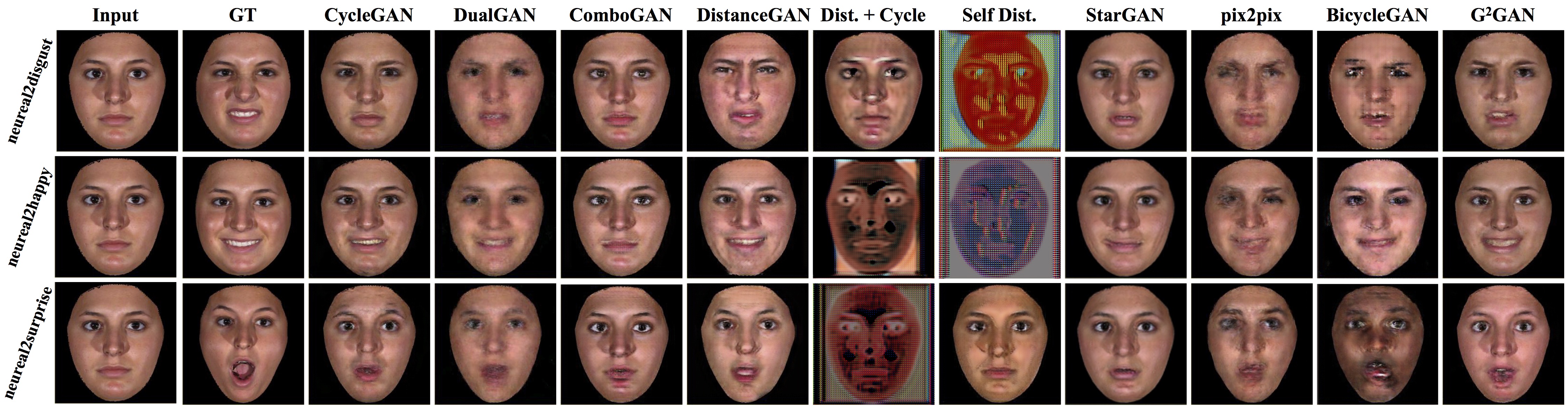}}
	\subfigure{\includegraphics[width=0.40\textwidth]{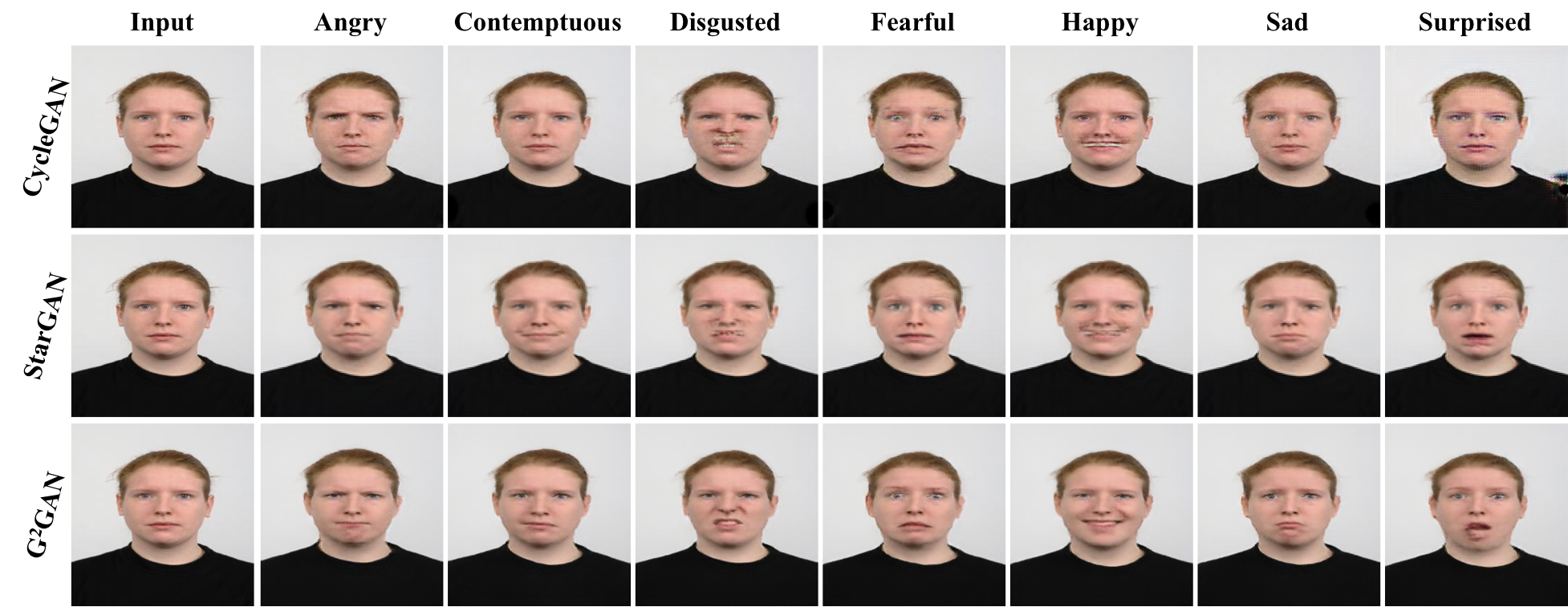}}
	\caption{Comparison with different models for facial expression translation on Bu3dfe (Left) and RaFD (Right) datasets.}
	\label{fig:comparison_bu3dfe}
\end{figure*}

\noindent\textbf{Task 2: Facial Expression Synthesis.}
We adopt three face datasets (\ie~AR, Bu3dfe and RaFD) for the facial expression synthesis task with similar settings as in StarGAN. Note that for AR dataset, we not only show the translation results of the neutral expression to other non-neutral expressions as in \cite{choi2017stargan}, but also present the opposite mappings, \ie~from non-neutral expressions to neutral expression. For Bu3dfe dataset, we only show the translation results from neutral expression to other non-neutral expressions as in \cite{choi2017stargan} because of the space limitation. As can be seen in Figure \ref{fig:comparison_ar}, Dist. + Cycle and Self Dist. fail to produce faces similar to the target domain. DualGAN generates reasonable but blurry faces. DistanceGAN, StarGAN, pix2pix and BicycleGAN produce much sharper results, but still contain some artifacts in the generated faces, \emph{e.g.}, twisted mouths of StarGAN, pix2pix and BicycleGAN on ``neutral2fear'' task. CycleGAN, ComboGAN and G$^2$GAN work better than other baselines on this dataset. 
We can also observe similar results on the Bu3dfe dataset as shown in Figure \ref{fig:comparison_bu3dfe} (Left).
Finally, we present results on the RaFD dataset in Figure~\ref{fig:comparison_bu3dfe} (Right). 
We can observe that our method achieves visually better results than CycleGAN and StarGAN.

\begin{figure}[!t] \tiny
	\centering
	\includegraphics[width=0.75\linewidth]{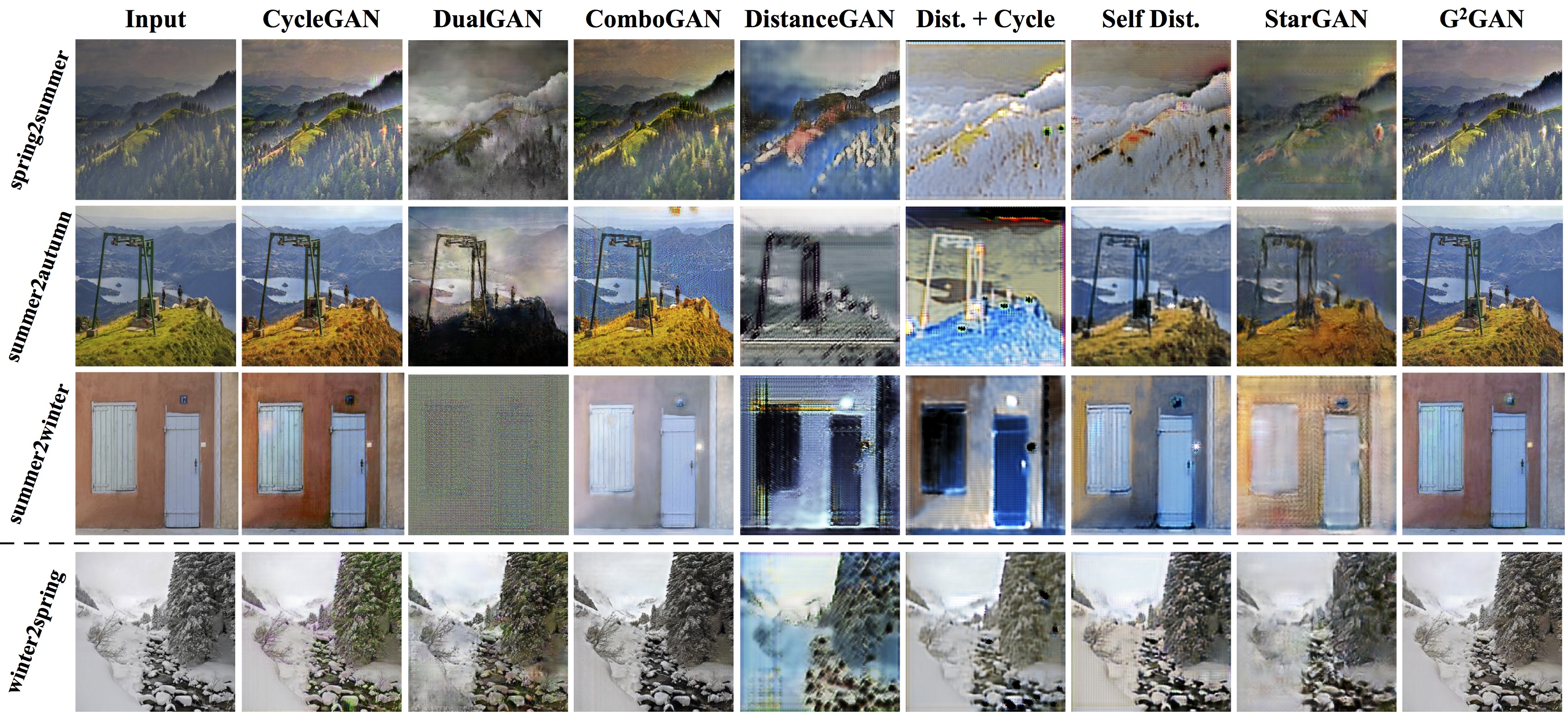}
	\caption{Comparison with different models for season translation on Alps. 
	}
	\label{fig:comparison_alps}
\end{figure}

\noindent\textbf{Task 3: Season Translation.}
We also validate G$^2$GAN on the season translation task.
The qualitative results are illustrated in Figure \ref{fig:comparison_alps}. Note that we did not show pix2pix and BicycleGAN results on Alps dataset since this dataset does not contain ground-truth images to train these two models. Obviously DualGAN, DistanceGAN, Dist. + Cycle, Self Dist. fail to produce reasonable results.
StarGAN produces reasonable but blurry results, and there are some artifacts in the generated images.
CycleGAN, ComboGAN and the proposed G$^2$GAN are able to generate better results than other baselines. However, ComboGAN yields some artifacts in some cases, such as the ``summer2autumn'' sub-task.
We also present one failure case of our method on this dataset in the last row of Figure \ref{fig:comparison_alps}. 
Our method produces images similar to the input domain, while CycleGAN and DualGAN generate visually better results compared with G$^2$GAN on ``winter2spring'' sub-task.
It is worth noting that CycleGAN and DualGAN need to train twelve generators on this dataset, while G$^2$GAN only requires two generators, and thus our model complexity is significantly lower.

\begin{figure}[!t] \tiny
	\begin{floatrow}
		\capbtabbox{%
			\resizebox{0.78\linewidth}{!} {
				\begin{tabular}{l|c|c|c} \toprule
					Model                       			          & AMT          & IS             & FID \\ \midrule		
					CycleGAN  \cite{zhu2017unpaired} & 19.5         &  1.6942     &   52.8230 \\ \hline
					StarGAN \cite{choi2017stargan}     & 24.7         &   1.6695     &  51.6929\\ \hline 
					G$^2$GAN (Ours)                          & \textbf{29.1}& \textbf{1.7187}   & \textbf{51.2765 }\\ \bottomrule 	
			\end{tabular}}
		}{
			\caption{Results on RaFD.}
			\label{tab:rafd}
		}
		\capbtabbox{
			\resizebox{0.9\linewidth}{!}{%
				\begin{tabular}{l|c|c|c} \toprule
					Model                            &  AMT                                  &  FID                     & CA \\ \midrule
					CycleGAN \cite{zhu2017unpaired} (ICCV 2017)  &  16.8\%$\pm$1.9\%              &  47.4823              & 73.72\% \\ \hline
					StarGAN \cite{choi2017stargan} (CVPR 2018) &  13.9\%$\pm$1.4\%             &  58.1562               & 44.63\% \\ \hline
					G$^2$GAN (Ours) &  \textbf{19.8\%$\pm$2.4\%}    &  \textbf{43.7473} &  \textbf{78.84\%}  \\ \hline
					Real Data   &  -                                       &  -                         & 91.34\% \\ \bottomrule		
			\end{tabular}}
		}{
			\caption{Results on collection style set.}
			\label{tab:result_paint}
		}
	\end{floatrow}
\end{figure}

\begin{figure}[!t]\tiny
	\centering
	\includegraphics[width=0.76\linewidth]{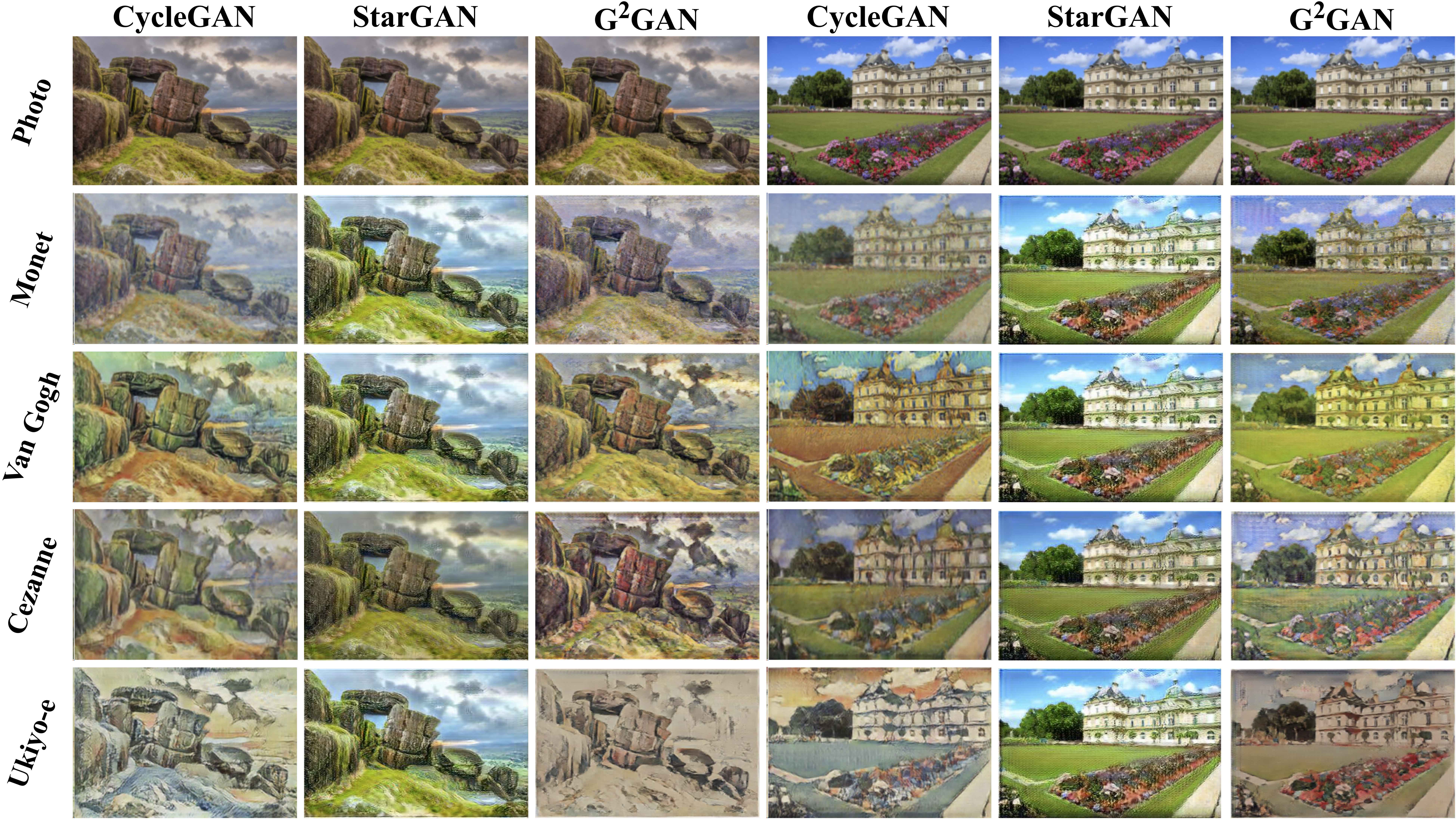}
	\caption{Comparison on the multi-domain painting style transfer task. }
	\label{fig:style_results}
\end{figure}

\noindent\textbf{Task 4: Painting Style Transfer.}
Figure~\ref{fig:style_results} shows the comparison results on the painting style dataset with CycleGAN and StarGAN. 
We observe that StarGAN produces less diverse generations crossing different styles compared with CycleGAN and G$^2$GAN. 
G$^2$GAN has comparable performance with CycleGAN, requiring only one single model for all the styles, and thus the network complexity is remarkably lower compared with CycleGAN which trains an individual model for each pair of styles.

\noindent\textbf{Quantitative Comparison on All Tasks.} We also provide quantitative results on the four tasks. Different metrics are considered including: (i) AMT perceptual studies~\cite{zhu2017unpaired,isola2017image}, (ii) Inception Score (IS) \cite{salimans2016improved}, (iii) Fr\'echet Inception Distance (FID) \cite{heusel2017gans} and (iv) Classification Accuracy (CA) \cite{choi2017stargan}. We follow the same perceptual study protocol from CycleGAN and StarGAN. Tables \ref{tab:rafd}, \ref{tab:result_paint} and \ref{tab:result_amt} report the performance of the AMT perceptual test, which is a ``real vs fake" perceptual metric assessing the realism from a holistic aspect. For Facades dataset, we split it into two subtasks as in \cite{zhu2017unpaired}, label$\rightarrow$photo and photo$\rightarrow$label. For the other datasets, we report the average performance of all mappings. 
Note that from Tables \ref{tab:rafd}, \ref{tab:result_paint} and \ref{tab:result_amt}, the proposed G$^2$GAN achieves very competitive results compared with the other baselines. Note that G$^2$GAN significantly outperforms StarGAN trained using one generator on most of the metrics and on all the datasets.
Note that paired pix2pix shows worse results than unpaired methods in Table \ref{tab:result_amt}, which can be also observed in DualGAN \cite{yi2017dualgan}.

We also use the Inception Score (IS) \cite{salimans2016improved} to measure the quality of generated images. 
Tables \ref{tab:rafd} and \ref{tab:result} report the results. 
As discussed before, the proposed G$^2$GAN generates sharper, more photo-realistic and reasonable results than Dist. + Cycle, Self Dist. and StarGAN, while the latter models present slightly higher IS. However, higher IS does not necessarily mean higher image quality. High quality images may have small IS as demonstrated in other image generation \cite{ma2017pose} and super-resolution works \cite{johnson2016perceptual,shi2016real}. 
Moreover, we employ FID \cite{heusel2017gans} to measure the performance on RaFD and painting style datasets.
Results are shown in Tables \ref{tab:rafd} and  \ref{tab:result_paint}, we observe that G$^2$GAN achieves the best results compared with StarGAN and CycleGAN.

\begin{table*}[!t] \tiny
	\centering
	\caption{AMT ``real vs fake'' study on Facades, AR, Alps, Bu3dfe datasets.}
	\resizebox{0.75\linewidth}{!}{%
		\begin{tabular}{l|c|c|c|c|c} \toprule
			\% Turkers label \textit{real}                  & label$\rightarrow$photo  & photo$\rightarrow$label   & AR                       & Alps               & Bu3dfe                     \\ \midrule
			
			CycleGAN \cite{zhu2017unpaired} (ICCV 2017)     & 8.8\% $\pm$ 1.5\%        & 4.8\% $\pm$ 0.8\%         & \textbf{24.3\%$\pm$1.7\%}& 39.6\% $\pm$ 1.4\% & 16.9\% $\pm$ 1.2\%         \\ \hline
			DualGAN \cite{yi2017dualgan} (ICCV 2017)        & 0.6\% $\pm$ 0.2\%        & 0.8\% $\pm$ 0.3\%         & 1.9\% $\pm$ 0.6\%        & 18.2\% $\pm$ 1.8\% & 3.2\% $\pm$ 0.4\%          \\ \hline
			ComboGAN \cite{anoosheh2017combogan} (CVPR 2018)& 4.1\% $\pm$ 0.5\%        & 0.2\% $\pm$ 0.1\%         & 4.7\% $\pm$ 0.9\%        & 34.3\% $\pm$ 2.2\% & \textbf{25.3\% $\pm$ 1.6\%}\\ \hline
			DistanceGAN \cite{benaim2017one} (NIPS 2017)    & 5.7\% $\pm$ 1.1\%        & 1.2\% $\pm$ 0.5\%         & 2.7\% $\pm$ 0.7\%        & 4.4\% $\pm$ 0.3\%  & 6.5\% $\pm$ 0.7\%          \\ \hline 
			Dist. + Cycle \cite{benaim2017one} (NIPS 2017)  & 0.3\% $\pm$ 0.2\%        & 0.2\% $\pm$ 0.1\%         & 1.3\% $\pm$ 0.5\%        & 3.8\% $\pm$ 0.6\%  & 0.3\% $\pm$ 0.1\%          \\ \hline 
			Self Dist. \cite{benaim2017one} (NIPS 2017)     & 0.3\% $\pm$ 0.1\%        & 0.1\% $\pm$ 0.1\%         & 0.1\% $\pm$ 0.1\%        & 5.7\% $\pm$ 0.5\%  & 1.1\% $\pm$ 0.3\%          \\ \hline 
			StarGAN \cite{choi2017stargan} (CVPR 2018 )     & 3.5\% $\pm$ 0.7\%        & 1.3\% $\pm$ 0.3\%         & 4.1\% $\pm$ 1.3\%        & 8.6\% $\pm$ 0.7\%  & 9.3\% $\pm$ 0.9\%          \\ \hline
			pix2pix \cite{isola2017image} (CVPR 2017)       & 4.6\% $\pm$ 0.5\%        & 1.5\% $\pm$ 0.4\%         & 2.8\% $\pm$ 0.6\%        & -                  & 3.6\% $\pm$ 0.5\%          \\ \hline
			BicycleGAN \cite{zhu2017toward} (NIPS 2017)     & 5.4\% $\pm$ 0.6\%        & 1.1\% $\pm$ 0.3\%         & 2.1\% $\pm$ 0.5\%        & -                  & 2.7\% $\pm$ 0.4\%          \\ \hline\hline 		
			G$^2$GAN (Ours, fully-sharing)                  & 4.6\% $\pm$ 0.9\%        & 2.4\% $\pm$ 0.4\%         & 6.8\% $\pm$ 0.6\%        & 15.4\% $\pm$ 1.9\% & 13.1\% $\pm$ 1.3\%         \\ 
			G$^2$GAN (Ours, partially-sharing)              & 8.2\% $\pm$ 1.2\%        & 3.6\% $\pm$ 0.7\%         & 16.8\% $\pm$1.2\%        & 36.7\% $\pm$ 2.3\% & 18.9\% $\pm$ 1.1\%         \\
			G$^2$GAN (Ours, no-sharing)                     & \textbf{10.3\%$\pm$1.6\%}& \textbf{5.6\% $\pm$ 0.9\%}& 22.8\% $\pm$1.9\%        & \textbf{47.7\%$\pm$2.8\%} & 23.6\% $\pm$ 1.7\%  \\ \bottomrule		
	\end{tabular}}
	\label{tab:result_amt}
\end{table*}

Finally, we compute the Classification Accuracy (CA) on the synthesized images as in~\cite{choi2017stargan}. We train classifiers on the AR, Alps, Bu3dfe, Collection datasets respectively.
For each dataset, we take the real image as training data and the generated images of different models as testing data. 
The intuition behind this setting is that if the generated images are realistic and follow the distribution of the images in the target domain, the classifiers trained on real images will be able to classify the generated image correctly. 
For AR, Alps and Collection datasets we list top 1 accuracy, while for Bu3dfe we report top 1 and top 5 accuracy. 
Tables \ref{tab:result_paint}  and  \ref{tab:result} show the results. 
Note that G$^2$GAN outperforms the baselines on AR, Bu3dfe and Collection datasets. 
On the Alps dataset, StarGAN achieves slightly better performance than ours but the generated images by our model contains less artifacts than StarGAN as shown in Figure \ref{fig:comparison_alps}.

\subsection{Model Analysis}

\begin{table*}[!t] \tiny
	\centering
	\caption{Results of Inception Score (IS) and Classification Accuracy (CA). }
	\resizebox{0.76\linewidth}{!}{
		\begin{tabular}{l|c|c|c|c|c|c|c} \toprule
			\multirow{2}{*}{Model}                          & Facades          & \multicolumn{2}{c|}{AR}           & \multicolumn{2}{c|}{Alps}             & \multicolumn{2}{c}{Bu3dfe}   \\ \cline{2-8}
			& IS               & IS              & CA              & IS              & CA                  & IS              & CA  \\ \midrule
			CycleGAN \cite{zhu2017unpaired} (ICCV 2017)     & 3.6098           & 2.8321          & @1:27.333\%     & 4.1734          & @1:42.250\%         & 1.8173          & @1:48.292\%, @5:94.167\% \\ \hline
			DualGAN \cite{yi2017dualgan} (ICCV 2017)        & 3.7495           & 1.9148          & @1:28.667\%     & 4.2661          & @1:53.488\%         & 1.7176          & @1:40.000\%, @5:90.833\% \\ \hline
			ComboGAN \cite{anoosheh2017combogan} (CVPR 2018)& 3.1289           & 2.4750          & @1:28.250\%     & 4.2438          & @1:62.750\%         & 1.7887          & @1:40.459\%, @5:90.714\% \\ \hline
			DistanceGAN \cite{benaim2017one} (NIPS 2017)    & 3.9988           & 2.3455          & @1:26.000\%     & 4.8047          & @1:31.083\%         & 1.8974          & @1:46.458\%, @5:90.000\% \\ \hline 
			Dist. + Cycle \cite{benaim2017one} (NIPS 2017)  & 2.6897           & \textbf{3.5554} & @1:14.667\%     & \textbf{5.9531} & @1:29.000\%         & 3.4618          & @1:26.042\%, @5:79.167\% \\ \hline 
			Self Dist. \cite{benaim2017one} (NIPS 2017)     & 3.8155           & 2.1350          & @1:21.333\%     & 5.0584          & @1:34.917\%         & \textbf{3.4620} & @1:10.625\%, @5:74.167\% \\ \hline 
			StarGAN \cite{choi2017stargan} (CVPR 2018)     & \textbf{4.3182}  & 2.0290          & @1:26.250\%     & 3.3670          & @1:\textbf{65.375\%}& 1.5640          & @1:52.704\%, @5:94.898\% \\ \hline
			pix2pix \cite{isola2017image} (CVPR 2017)       & 3.6664           & 2.2849          & @1:22.667\%     & -               & -                   & 1.4575          & @1:44.667\%, @5:91.750\% \\ \hline
			BicycleGAN \cite{zhu2017toward} (NIPS 2017)     & 3.2217           & 2.0859          & @1:28.000\%     & -               & -                   & 1.7373          & @1:45.125\%, @5:93.125\% \\ \hline\hline
			G$^2$GAN (Ours, fully-sharing)                  & 4.2615           & 2.3875          & @1:28.396\%     & 3.6597          & @1:61.125\%         & 1.9728          & @1:52.985\%, @5:95.165\% \\
			G$^2$GAN (Ours, partially-sharing)              & 4.1689           & 2.4846          & @1:28.835\%     & 4.0158          & @1:62.325\%         & 1.5896          & @1:53.456\%, @5:95.846\% \\
			G$^2$GAN (Ours, no-sharing)                     & 4.0819           & 2.6522          & @1:\textbf{29.667\%} & 4.3773     & @1:63.667\%         & 1.8714          & @1:\textbf{55.625\%}, @5:\textbf{96.250\%}  \\		
			\bottomrule		
	\end{tabular}}
	\label{tab:result}
\end{table*}

\noindent\textbf{Model Component Analysis.}
We conduct an ablation study of the proposed G$^2$GAN on Facades, AR and Bu3dfe datasets. We show the results without the conditional identity preserving loss (I), multi-scale SSIM loss (S), color cycle-consistency loss (C) and double discriminators strategy (D), respectively. We also consider using two different discriminators as in~\cite{xu2017face,nguyen2017dual,tang2018gesturegan} to further boost our performance. To study the parameter sharing for the dual generator, we perform experiments on different schemes including: fully-sharing, \ie~the two generators share the same parameters, partially-sharing, \ie~only the encoder part shares the same parameters, no-sharing, \ie~two independent generators. The basic generator structure follows~\cite{choi2017stargan}. Quantitative results of the AMT score and the classification accuracy are reported in Table \ref{tab:Ablation}. Without using double discriminators slightly degrades performance, meaning that the proposed G$^2$GAN can achieve good results trained using the dual generator and one discriminator. However, removing the conditional identity preserving loss, multi-scale SSIM loss and color cycle-consistency loss substantially degrades the performance, meaning that the proposed joint optimization objectives are particularly important to stabilize the training and thus produce better generation results. For the parameter sharing, as shown in Table~\ref{tab:result_amt},  \ref{tab:result} and \ref{tab:computational}, we observe that different-level parameter sharing influences both the generation performance and the model capacity, demonstrating our initial motivation. 

\begin{table*}[!t] \tiny
	\centering
	\caption{Evaluation of different variants of G$^2$GAN on Facades, AR and Bu3dfe datasets. All: full version of G$^2$GAN, I: Identity preserving loss, S: multi-scale SSIM loss, C: Color cycle-consistency loss, D: Double discriminators strategy.}
	\resizebox{0.76\linewidth}{!}{
		\begin{tabular}{l|c|c|c|c|c|c} \toprule
			\multirow{2}{*}{Model}   & label$\rightarrow$photo        & photo$\rightarrow$label        & \multicolumn{2}{c|}{AR}                       & \multicolumn{2}{c}{Bu3dfe}   \\ \cline{2-7}
			& \% Turkers label \textit{real} & \% Turkers label \textit{real} & \% Turkers label \textit{real} & CA           & \% Turkers label \textit{real}   & CA  \\ \midrule	
			All                      & \textbf{10.3\% $\pm$ 1.6\%}    & \textbf{5.6\% $\pm$ 0.9\%}              & \textbf{22.8\% $\pm$1.9\% }             & @1:\textbf{29.667\%}  & \textbf{23.6\% $\pm$ 1.7\%}               & @1:\textbf{55.625\%}, @5:\textbf{96.250\%} \\ \hline
			All - I                  & 2.6\% $\pm$ 0.4\%              & 4.2\% $\pm$ 1.1\%              & 4.7\% $\pm$ 0.8\%              & @1:29.333\%  & 16.3\% $\pm$ 1.1\%               & @1:53.739\%, @5:95.625\%\\ \hline
			All - S - C              & 4.4\% $\pm$ 0.6\%              & 4.8\% $\pm$ 1.3\%              & 8.7\% $\pm$ 0.6\%              & @1:28.000\%  & 14.4\% $\pm$ 1.2\%               & @1:42.500\%, @5:95.417\%\\ \hline
			All - S - C - I          & 2.2\% $\pm$ 0.3\%              & 3.9\% $\pm$ 0.8\%              & 2.1\% $\pm$ 0.4\%              & @1:24.667\%  & 13.6\% $\pm$ 1.2\%               & @1:41.458\%, @5:95.208\%\\ \hline
			All - D                  & 9.0\% $\pm$ 1.5\%              & 5.3\% $\pm$ 1.1\%              & 21.7\% $\pm$ 1.7\%             & @1:28.367\%  & 22.3\% $\pm$ 1.6\%               & @1:53.375\%, @5:95.292\%\\ \hline
			All - D - S              & 3.3\% $\pm$ 0.7\%              & 4.5\% $\pm$ 1.1\%              & 14.7\% $\pm$1.7\%              & @1:27.333\%  & 20.1\% $\pm$ 1.4\%               & @1:42.917\%, @5:91.250\%\\ \hline
			All - D - C              & 8.7\% $\pm$ 1.3\%              & 5.1\% $\pm$ 0.9\%              & 19.4\% $\pm$1.5\%              & @1:28.000\%  & 21.6\% $\pm$ 1.4\%               & @1:45.833\%, @5:93.875\%\\ \bottomrule		
	\end{tabular}}
	\label{tab:Ablation}
\end{table*}

\begin{table*}[!t] \tiny
	\centering
	\caption{Comparison of the overall model capacity with different models.}
	\resizebox{0.52\linewidth}{!}{%
		\begin{tabular}{l|c|c} \toprule		    
			Method  & \# Models   & \# Parameters with $m$=7\\ \midrule
			\tht{l}{pix2pix \cite{isola2017image} (CVPR 2017) \\ BicycleGAN \cite{zhu2017toward} (NIPS 2017)}  & $A_m^2{=}m(m-1)$ & \tht{l}{57.2M $\times$ 42 \\ 64.3M $\times$ 42} \\ \hline
			\tht{l}{CycleGAN \cite{zhu2017unpaired} (ICCV 2017) \\ DiscoGAN \cite{kim2017learning} (ICML 2017) \\ DualGAN \cite{yi2017dualgan} (ICCV 2017) \\ DistanceGAN \cite{benaim2017one} (NIPS 2017)} & $C_m^2{=}\frac{m(m-1)}{2}$ & \tht{l}{
				52.6M $\times$ 21 \\ 16.6M $\times$ 21 \\ 178.7M $\times$ 21 \\ 
				52.6M $\times$ 21} \\ \hline
			ComboGAN \cite{anoosheh2017combogan} (CVPR 2018) & $m$ & 14.4M $\times$ 7\\ \hline
			StarGAN \cite{choi2017stargan} (CVPR 2018) & 1 & 53.2M $\times$ 1 \\ \hline\hline
			G$^2$GAN (Ours, fully-sharing)   & 1 & 53.2M $\times$ 1  \\
			G$^2$GAN (Ours, partial-sharing) & 1 & 53.8M $\times$ 1 \\
			G$^2$GAN (Ours, no-sharing)	     & 1 & 61.6M $\times$ 1 \\ \bottomrule		
	\end{tabular}}
	\label{tab:computational}
\end{table*}

\noindent\textbf{Overall Model Capacity Analysis.}
We compare the overall model capacity with other baselines.
The number of models and the number of model parameters on Bu3dfe dataset for different $m$ image domains are shown in Table~\ref{tab:computational}.
BicycleGAN and pix2pix are supervised models so that they need to train $A_m^2$ models for $m$ image domains.
CycleGAN, DiscoGAN, DualGAN, DistanceGAN are unsupervised methods, and they require $C_m^2$ models to learn $m$ image domains, but each of them contains two generators and two discriminators.
ComboGAN requires only $m$ models to learn all the mappings of $m$ domains, while StarGAN and G$^2$GAN only need to train one model to learn all the mappings of $m$ domains. We also report the number of parameters on Bu3dfe dataset, this dataset contains 7 different expressions, which means $m{=}7$. Note that DualGAN uses fully connected layers in the generators, which brings significantly larger number of parameters. CycleGAN and DistanceGAN have the same architectures, which means they have the same number of parameters. Moreover, G$^2$GAN uses less parameters compared with the other baselines except StarGAN, but we achieve significantly better generation performance in most metrics as shown in Tables \ref{tab:rafd}, \ref{tab:result_paint}, \ref{tab:result_amt} and  \ref{tab:result}. 
When we employ the parameter sharing scheme, our performance is only slightly lower (still outperforming StarGAN) while the number of parameters is comparable with StarGAN.   

\section{Conclusion}

We propose a novel Dual Generator Generative Adversarial Network (G$^2$GAN), a robust and scalable generative model that allows performing unpaired image-to-image translation for multiple domains using only dual generators within a single model.
The dual generators, allowing for different network structures and different-level parameter sharing, are designed for the translation and the reconstruction tasks.
Moreover, we explore jointly using different loss functions to optimize the proposed G$^2$GAN, and thus generating images with high quality. 
Extensive experiments on different scenarios demonstrate that the proposed G$^2$GAN achieves more photo-realistic results and less model capacity than other baselines.
In the future, we will focus on the face aging task~\cite{wang2018recurrent}, which aims to generate facial image with different ages in a continuum.

\clearpage
\small
\bibliographystyle{splncs04}
\bibliography{egbib}

\begin{thebibliography}{10}
\providecommand{\url}[1]{\texttt{#1}}
\providecommand{\urlprefix}{URL }
\providecommand{\doi}[1]{https://doi.org/#1}

\bibitem{anoosheh2017combogan}
Anoosheh, A., Agustsson, E., Timofte, R., Van~Gool, L.: Combogan: Unrestrained
  scalability for image domain translation. In: CVPR Workshop (2018)

\bibitem{arjovsky2017wasserstein}
Arjovsky, M., Chintala, S., Bottou, L.: Wasserstein gan. In: ICML (2017)

\bibitem{benaim2017one}
Benaim, S., Wolf, L.: One-sided unsupervised domain mapping. In: NIPS (2017)

\bibitem{brock2016neural}
Brock, A., Lim, T., Ritchie, J.M., Weston, N.: Neural photo editing with
  introspective adversarial networks. In: ICLR (2017)

\bibitem{choi2017stargan}
Choi, Y., Choi, M., Kim, M., Ha, J.W., Kim, S., Choo, J.: Stargan: Unified
  generative adversarial networks for multi-domain image-to-image translation.
  In: CVPR (2018)

\bibitem{goodfellow2014generative}
Goodfellow, I., Pouget-Abadie, J., Mirza, M., Xu, B., Warde-Farley, D., Ozair,
  S., Courville, A., Bengio, Y.: Generative adversarial nets. In: NIPS (2014)

\bibitem{heusel2017gans}
Heusel, M., Ramsauer, H., Unterthiner, T., Nessler, B., Hochreiter, S.: Gans
  trained by a two time-scale update rule converge to a local nash equilibrium.
  In: NIPS (2017)

\bibitem{isola2017image}
Isola, P., Zhu, J.Y., Zhou, T., Efros, A.A.: Image-to-image translation with
  conditional adversarial networks. In: CVPR (2017)

\bibitem{johnson2016perceptual}
Johnson, J., Alahi, A., Fei-Fei, L.: Perceptual losses for real-time style
  transfer and super-resolution. In: ECCV (2016)

\bibitem{kim2017learning}
Kim, T., Cha, M., Kim, H., Lee, J., Kim, J.: Learning to discover cross-domain
  relations with generative adversarial networks. In: ICML (2017)

\bibitem{kingma2014adam}
Kingma, D., Ba, J.: Adam: A method for stochastic optimization. In: ICLR (2015)

\bibitem{langner2010presentation}
Langner, O., Dotsch, R., Bijlstra, G., Wigboldus, D.H., Hawk, S.T.,
  Van~Knippenberg, A.: Presentation and validation of the radboud faces
  database. Cognition and emotion  \textbf{24}(8),  1377--1388 (2010)

\bibitem{li2016precomputed}
Li, C., Wand, M.: Precomputed real-time texture synthesis with markovian
  generative adversarial networks. In: ECCV (2016)

\bibitem{li2018beautygan}
Li, T., Qian, R., Dong, C., Liu, S., Yan, Q., Zhu, W., Lin, L.: Beautygan:
  Instance-level facial makeup transfer with deep generative adversarial
  network. In: ACM MM (2018)

\bibitem{li2017generative}
Li, Y., Liu, S., Yang, J., Yang, M.H.: Generative face completion. In: CVPR
  (2017)

\bibitem{liang2017generative}
Liang, X., Zhang, H., Xing, E.P.: Generative semantic manipulation with
  contrasting gan. In: ECCV (2018)

\bibitem{liu2017unsupervised}
Liu, M.Y., Breuel, T., Kautz, J.: Unsupervised image-to-image translation
  networks. In: NIPS (2017)

\bibitem{liu2016coupled}
Liu, M.Y., Tuzel, O.: Coupled generative adversarial networks. In: NIPS (2016)

\bibitem{ma2017pose}
Ma, L., Jia, X., Sun, Q., Schiele, B., Tuytelaars, T., Van~Gool, L.: Pose
  guided person image generation. In: NIPS (2017)

\bibitem{mansimov2015generating}
Mansimov, E., Parisotto, E., Ba, J.L., Salakhutdinov, R.: Generating images
  from captions with attention. In: ICLR (2015)

\bibitem{mao2017least}
Mao, X., Li, Q., Xie, H., Lau, R.Y., Wang, Z., Smolley, S.P.: Least squares
  generative adversarial networks. In: ICCV (2017)

\bibitem{martinez1998ar}
Martinez, A.M.: The ar face database. CVC TR  (1998)

\bibitem{mirza2014conditional}
Mirza, M., Osindero, S.: Conditional generative adversarial nets. arXiv
  preprint arXiv:1411.1784  (2014)

\bibitem{nguyen2017dual}
Nguyen, T., Le, T., Vu, H., Phung, D.: Dual discriminator generative
  adversarial nets. In: NIPS (2017)

\bibitem{odena2016conditional}
Odena, A., Olah, C., Shlens, J.: Conditional image synthesis with auxiliary
  classifier gans. In: ICML (2017)

\bibitem{park2017transformation}
Park, E., Yang, J., Yumer, E., Ceylan, D., Berg, A.C.: Transformation-grounded
  image generation network for novel 3d view synthesis. In: CVPR (2017)

\bibitem{perarnau2016invertible}
Perarnau, G., van~de Weijer, J., Raducanu, B., {\'A}lvarez, J.M.: Invertible
  conditional gans for image editing. In: NIPS Workshop (2016)

\bibitem{qi2017loss}
Qi, G.J.: Loss-sensitive generative adversarial networks on lipschitz
  densities. arXiv preprint arXiv:1701.06264  (2017)

\bibitem{reed2016generative}
Reed, S., Akata, Z., Yan, X., Logeswaran, L., Schiele, B., Lee, H.: Generative
  adversarial text-to-image synthesis. In: ICML (2016)

\bibitem{reed2016learning}
Reed, S.E., Akata, Z., Mohan, S., Tenka, S., Schiele, B., Lee, H.: Learning
  what and where to draw. In: NIPS (2016)

\bibitem{ruder2017overview}
Ruder, S.: An overview of multi-task learning in deep neural networks. arXiv
  preprint arXiv:1706.05098  (2017)

\bibitem{salimans2016improved}
Salimans, T., Goodfellow, I., Zaremba, W., Cheung, V., Radford, A., Chen, X.:
  Improved techniques for training gans. In: NIPS (2016)

\bibitem{sangkloy2016scribbler}
Sangkloy, P., Lu, J., Fang, C., Yu, F., Hays, J.: Scribbler: Controlling deep
  image synthesis with sketch and color. In: CVPR (2017)

\bibitem{shi2016real}
Shi, W., Caballero, J., Husz{\'a}r, F., Totz, J., Aitken, A.P., Bishop, R.,
  Rueckert, D., Wang, Z.: Real-time single image and video super-resolution
  using an efficient sub-pixel convolutional neural network. In: CVPR (2016)

\bibitem{shrivastava2016learning}
Shrivastava, A., Pfister, T., Tuzel, O., Susskind, J., Wang, W., Webb, R.:
  Learning from simulated and unsupervised images through adversarial training.
  In: CVPR (2017)

\bibitem{shrivastava2017learning}
Shrivastava, A., Pfister, T., Tuzel, O., Susskind, J., Wang, W., Webb, R.:
  Learning from simulated and unsupervised images through adversarial training.
  In: CVPR (2017)

\bibitem{shu2017neural}
Shu, Z., Yumer, E., Hadap, S., Sunkavalli, K., Shechtman, E., Samaras, D.:
  Neural face editing with intrinsic image disentangling. In: CVPR (2017)

\bibitem{siarohin2018deformable}
Siarohin, A., Sangineto, E., Lathuili{\`e}re, S., Sebe, N.: Deformable gans for
  pose-based human image generation. In: CVPR (2018)

\bibitem{taigman2016unsupervised}
Taigman, Y., Polyak, A., Wolf, L.: Unsupervised cross-domain image generation.
  In: ICLR (2017)

\bibitem{tang2018gesturegan}
Tang, H., Wang, W., Xu, D., Yan, Y., Sebe, N.: Gesturegan for hand
  gesture-to-gesture translation in the wild. In: ACM MM (2018)

\bibitem{tylevcek2013spatial}
Tyle{\v{c}}ek, R., {\v{S}}{\'a}ra, R.: Spatial pattern templates for
  recognition of objects with regular structure. In: GCPR (2013)

\bibitem{wang2018every}
Wang, W., Alameda-Pineda, X., Xu, D., Fua, P., Ricci, E., Sebe, N.: Every smile
  is unique: Landmark-guided diverse smile generation. In: CVPR (2018)

\bibitem{wang2018recurrent}
Wang, W., Yan, Y., Cui, Z., Feng, J., Yan, S., Sebe, N.: Recurrent face aging
  with hierarchical autoregressive memory. IEEE TPAMI  (2018)

\bibitem{wang2004image}
Wang, Z., Bovik, A.C., Sheikh, H.R., Simoncelli, E.P.: Image quality
  assessment: from error visibility to structural similarity. IEEE TIP
  \textbf{13}(4),  600--612 (2004)

\bibitem{wang2003multiscale}
Wang, Z., Simoncelli, E.P., Bovik, A.C.: Multiscale structural similarity for
  image quality assessment. In: Asilomar Conference on Signals, Systems and
  Computers (2003)

\bibitem{xu2017face}
Xu, R., Zhou, Z., Zhang, W., Yu, Y.: Face transfer with generative adversarial
  network. arXiv preprint arXiv:1710.06090  (2017)

\bibitem{yeh2016semantic}
Yeh, R., Chen, C., Lim, T.Y., Hasegawa-Johnson, M., Do, M.N.: Semantic image
  inpainting with perceptual and contextual losses. In: CVPR (2017)

\bibitem{yi2017dualgan}
Yi, Z., Zhang, H., Gong, P.T., et~al.: Dualgan: Unsupervised dual learning for
  image-to-image translation. In: ICCV (2017)

\bibitem{yin20063d}
Yin, L., Wei, X., Sun, Y., Wang, J., Rosato, M.J.: A 3d facial expression
  database for facial behavior research. In: FGR (2006)

\bibitem{zhu2017unpaired}
Zhu, J.Y., Park, T., Isola, P., Efros, A.A.: Unpaired image-to-image
  translation using cycle-consistent adversarial networks. In: ICCV (2017)

\bibitem{zhu2017toward}
Zhu, J.Y., Zhang, R., Pathak, D., Darrell, T., Efros, A.A., Wang, O.,
  Shechtman, E.: Toward multimodal image-to-image translation. In: NIPS (2017)

\end{thebibliography}
\end{document}